%% file: main.tex
\documentclass[runningheads]{llncs}

\usepackage[T1]{fontenc}
\usepackage{graphicx}
\usepackage{url}
\usepackage{booktabs}
\usepackage{float}
\usepackage{amsmath,amssymb}
\usepackage{microtype}

\begin{document}

\title{Client Exits - A Way to Increase LLM Transparency on the Client Side}

\author{Marc Jansen}

\institute{Computer Science Institute\\University of Applied Sciences Ruhr West\\Bottrop\\Germany\\
\email{marc.jansen@hs-ruhrwest.de}}

\maketitle

\begin{abstract}
Textual recommendations generated by large language models (LLMs) can be difficult to inspect. This paper introduces \emph{Client Exits}, a lightweight mechanism for enriching LLM responses while separating semantic intent from client-specific presentation. The model inserts predefined, machine-interpretable markers into its output, and the receiving client decides whether and how to realize the corresponding enrichments. This allows supplementary information to be presented as graphics, text, speech, or other modalities without prescribing a concrete interface. We implement a prototype combining a finetuned Qwen model with a modified LibreChat client. Using the Question--Option--Criteria (QOC) approach including three XML-based client exits. We evaluate two QOC-oriented model variants trained on essentially the same examples, with client-exit annotations added for one variant. Across 100 test questions, the client-exit model includes all three markers in every response. An LLM judge, informed about the mechanism without being instructed to favor it, prefers this model's answers in 66\% of the comparisons. These results demonstrate successful marker generation on the evaluated test set and support the feasibility of client-controlled enrichment.

\end{abstract}

\input{sections/introduction/introduction}
\input{sections/state-of-the-art/state-of-the-art}
\input{sections/architecture/architecture}
\input{sections/implementation/implementation}
\input{sections/evaluation/evaluation}
\input{sections/conclusion-and-future-work/conclusion-and-future-work}

\bibliographystyle{splncs04}
\bibliography{references}

\end{document}

%% file: sections/introduction/introduction.tex
\section{Introduction}
\label{sec:introduction}
Large language models (LLMs) can formulate recommendations and explain them in natural language. For questions involving multiple alternatives and competing criteria, however, a textual answer alone may make it difficult to inspect how the alternatives compare, which criteria distinguish them, or whether other options are similarly suitable. Graphical overviews and statistical summaries offer complementary ways of presenting this information. The challenge addressed in this paper is how an LLM can indicate where such enrichments would be useful without prescribing how a particular client must present them.

This challenge concerns the boundary between answer generation and client-side presentation. Software engineering provides a useful precedent in predefined extension points, such as User Exits, which allow application-specific functionality to be introduced without modifying a standard implementation \cite{brehm2001tailoring}. At the presentation boundary of an LLM system, a corresponding mechanism should allow the producer to indicate an opportunity for enrichment while leaving its realization to the receiving client. In particular, the intended information should remain separate from its graphical, textual, or spoken representation.

We propose \emph{Client Exits} as a lightweight mechanism for this separation. An LLM inserts predefined, machine-interpretable markers at selected positions within its response. Before displaying the answer, the client identifies these markers and decides whether and how to integrate supplementary information. A marker may indicate an overview, an analysis, or an explanation, but does not specify a concrete interface layout. The same response can therefore support different presentations depending on the client's capabilities and purpose. The underlying research question is whether an LLM can learn to generate these semantic extension points reliably and whether responses from a model trained to include them are preferred over those from a comparable model without them.

This paper contributes the client-exit architecture, an end-to-end prototype, and an initial comparative evaluation. The prototype combines a finetuned Qwen model with a modified LibreChat client and uses the Question--Option--Criteria (QOC) approach to structure decision-oriented answers \cite{maclean2020questions,jansen2025increasing}. Three XML-based client exits support a graph of the question, options, and criteria; a ranking of criteria by their discriminating influence; and a comparison of alternatives with scores close to the winning score. These enrichments make the generated analysis available for inspection alongside the recommendation, rather than claiming to expose the model's internal computations.

For the evaluation, we finetuned two QOC-oriented model variants using essentially the same training examples, with client-exit markers added for one variant. Across 100 test questions, the client-exit model included all three markers in every response. An LLM judge, given a theoretical explanation of the mechanism without instructions to favor it, preferred this model's answer in 66\% of the comparisons. These findings demonstrate successful marker generation on the evaluated test set and a majority preference for the corresponding answers. They do not, however, establish improvements in factual accuracy or human understanding. The approach thus provides an implemented basis for client-controlled enrichment, with its practical transparency benefits remaining a subject for further evaluation.

%% file: sections/state-of-the-art/state-of-the-art.tex
\section{State of the Art}
\label{sec:state-of-the-art}
Extensibility is a long-established principle in software engineering. Rather than modifying a standard implementation whenever application-specific functionality is required, software systems can expose predefined extension points at which additional functionality may be introduced. A prominent example is the concept of \texttt{User Exits} in enterprise resource planning (ERP) systems. Brehm et al. \cite{brehm2001tailoring} identify User Exits as a distinct ERP-tailoring mechanism in which additional software code is introduced through predefined interfaces instead of modifying the underlying package. Similarly, Ng \cite{ng2001decision} discusses SAP R/3 Customer Exits as a mechanism for adding customer-specific functionality while preserving the maintainability and upgradeability of the standard system.

Thus, the provider determines where extension is permitted and specifies the corresponding interface, whereas the consumer determines whether and how the extension is realized.

A structurally related development can be observed in contemporary LLM systems. Increasingly, LLMs are embedded into larger systems in which functionality is distributed among models, external tools, data sources, and autonomous agents. The Model Context Protocol (MCP) standardizes access from LLM applications to external tools, resources, and contextual information \cite{hou2025model}. Agent2Agent (A2A), in contrast, addresses interoperability among autonomous agents and supports capability discovery, task delegation, message exchange, and the communication of resulting artifacts \cite{duan2025agent}. Recent taxonomies consequently distinguish agent-to-context communication, exemplified by MCP, from agent-to-agent communication, exemplified by A2A \cite{sander2026technical}.

MCP and A2A are not User Exits in the strict technical sense. Nevertheless, they serve the same broader architectural objective: functionality is composed through explicitly defined interfaces rather than being incorporated into a monolithic implementation. 

These mechanisms primarily extend what an LLM-based system can access, execute, or delegate. They do not directly address another relevant boundary: the transition from the semantic result generated by an LLM to its presentation by the receiving client.

The in this paper proposed \texttt{Client Exit} transfers the extension-point principle to this boundary. The LLM may insert a predefined, machine-interpretable marker into or alongside its response to indicate that an additional presentation or information artifact could be useful. The model determines the location and semantic purpose of the exit, while the receiving client retains authority over whether the marker is interpreted and how the corresponding enrichment is realized.

This constitutes an inversion of the classical User Exit. Whereas a User Exit allows externally provided code to participate in the execution of the standard system, a Client Exit does not execute client-supplied functionality. Instead, the producer exposes a semantic extension point and leaves its realization entirely to the consumer. Consequently, Client Exits are distinct from tool calls: a tool call requests an operation or capability, whereas a Client Exit specifies an optional enrichment of the presentation of a result.

A central characteristic of this approach is the separation of semantic content and representation. An enrichment may, for example, request an overview, statistical analysis, additional explanation, or evidence summary. Such semantic intents do not determine whether the information must be represented as text, graphics, speech, or another modality. Different clients may therefore realize the same Client Exit differently according to their capabilities, interaction modality, or local policy. An overview may be rendered graphically by a desktop client, textually by a terminal, or acoustically by a voice-oriented interface. In this respect, Client Exits implement a presentation-intent layer rather than prescribing a concrete View.

Recent work on generative and agent-driven user interfaces provides the closest related approaches. A2UI allows agents to communicate declarative interface descriptions that are rendered using client-controlled components \cite{zhou2026externalization}. MCP Apps similarly enables MCP tools to provide interactive user-interface resources to supporting hosts \cite{oo2026comparing}. Both approaches demonstrate the increasing importance of standardized interaction between agent-generated information and client-side presentation. Their abstraction level, however, differs from that proposed here. MCP Apps primarily provides an interactive UI artifact, while A2UI communicates declarative UI structures and components. Client Exits instead operate at the level of semantic presentation intent: they specify what additional information or presentation function would be useful without prescribing the concrete interface structure used to realize it.

The resulting research gap concerns a lightweight mechanism that combines predefined semantic extension points in LLM output with strict separation between enrichment semantics and representation, while leaving interpretation and realization under client control. Client Exits address this gap by transferring the established software-engineering principle of predefined extension points from application execution to the presentation boundary of LLM systems. The resulting architecture separates generation of the semantic result from its client-specific enrichment and representation.

%% file: sections/architecture/architecture.tex
\section{Architecture}
\label{sec:architecture}
The proposed architecture, as shown in Fig. \ref{fig:architecture} separates the generation of an answer from the presentation of supplementary information within that answer. A Large Language Model, optionally finetuned for this purpose, inserts special markers into its textual output. These markers act as \emph{client exits}: designated points at which the receiving client can enrich the answer with additional content.
The LLM determines where to place these markers, whereas the client determines whether and how to interpret and present them. Consequently, the generated response need not prescribe a particular user interface or output modality.

The interaction begins when a user enters a prompt into an LLM client. The client forwards the prompt to the LLM, which generates a response containing ordinary text and, where the model considers them appropriate, special markers. Markers can occur at different positions throughout the response, allowing supplementary information to be associated with the relevant parts of the answer rather than being confined to a separate, fixed interface element.
\begin{figure}[htb]
    \centering
    \includegraphics[width=\linewidth]{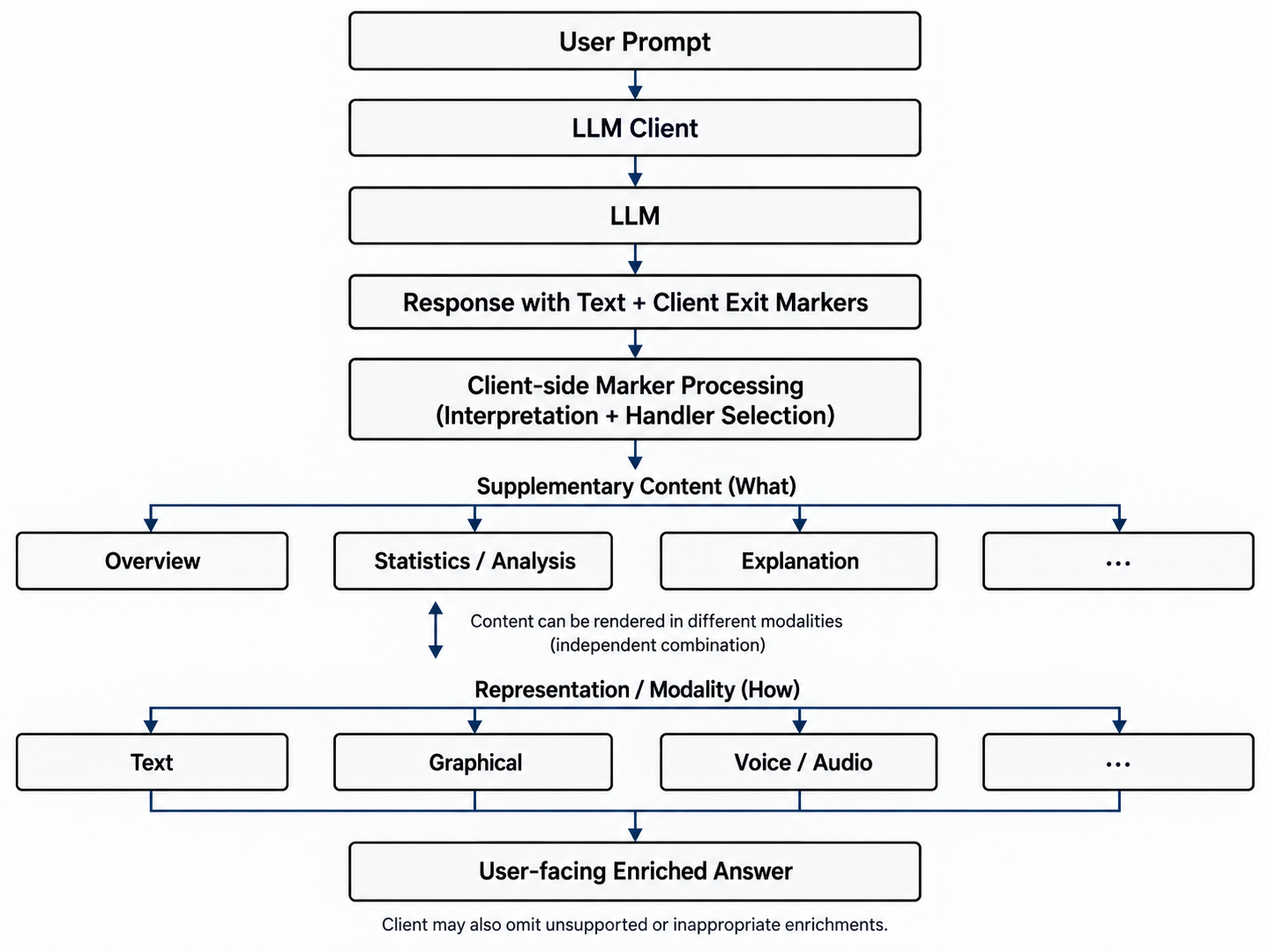}
    \caption{Client-exit architecture: the LLM inserts markers into its response, and the client interprets these markers to integrate supplementary information and adapt the presentation of the response.}
    \label{fig:architecture}
\end{figure}

Upon receiving the response, the client processes it before displaying it to the user. It identifies the markers, interprets their meaning, and determines how to handle each occurrence. For a supported marker, the client can obtain or compute supplementary content and integrate its presentation at the indicated position. The user therefore receives an enriched answer rather than the raw marker annotated output. This processing stage establishes a clear division of responsibility: the LLM proposes points for enrichment, while the client retains control over the resulting user experience.

The supplementary content may include visualizations, statistical analyses, explanations, or other information that the client considers useful in context. The marker provides an integration point; the corresponding client-side handler determines the concrete enrichment. To support this interaction, the model and the client need a shared marker convention that makes markers distinguishable from ordinary text and defines their intended meaning. Any information needed by a handler must be available through the marker, the surrounding response, or the client's context. The concrete marker syntax and handler implementation are separate from this architectural separation of responsibilities.

Presentation depends on the client's purpose and capabilities. A graphical client might render a chart or an expandable explanation, whereas a voice-based client might provide a spoken summary or an additional verbal explanation at the corresponding point in the answer. An audio-oriented client could instead use sound to convey supplementary information. Thus, the same marker-annotated response can support different presentations without requiring the LLM to generate a separate answer for each modality. The client can also choose not to render an enrichment when it is unsupported or inappropriate for the current interaction.

This separation is conceptually comparable to the Model--View--Controller (MVC) design pattern. In this analogy, the generated text, its markers, and the associated information constitute the data to be presented; client-side marker interpretation and handler selection perform a controller-like role; and the graphical, spoken, or other output constitutes the view. Here, the MVC notion of a model refers to the underlying data rather than to the LLM itself. The central similarity is that a common representation can give rise to differentviews through client-side decisions. The architecture adopts this separation without requiring every client to implement a strict MVC structure.

%% file: sections/implementation/implementation.tex
\section{Implementation}
\label{sec:implementation}
To demonstrate the architecture described in Section~\ref{sec:architecture}, we implemented a prototype combining a finetuned LLM with a modified LLM client. We first finetuned a Qwen3.8-27B model to follow the Question--Option--Criteria (QOC) thinking strategy \cite{maclean2020questions,jansen2025increasing} and to insert client exits into its responses. The finetuning thus addressed both the strategy used to structure the answer and the inclusion of markers that enable client-side enrichment.

In the prototype, client exits were represented as self-closing XML tags with an \texttt{exit} attribute identifying the requested client exit. The general form was:
\begin{verbatim}
<client exit="<name>" />
\end{verbatim}
Here, \texttt{<name>} denotes a placeholder for the client-exit identifier, not a literal part of the emitted attribute value. Identifiers can include a namespace prefix, allowing client exits to be grouped by thinking strategy. This naming convention enables a client to distinguish strategy-specific handlers and interpret exits differently for different strategies. The namespace is encoded in the attribute value rather than through an XML namespace declaration.

For the QOC strategy, we implemented the following three client exits:
\begin{verbatim}
<client exit="qoc.GraphOverview" />
<client exit="qoc.CriteriaInfluence" />
<client exit="qoc.SimilarGoodOptions" />
\end{verbatim}
The shared \texttt{qoc} prefix associates these exits with the QOC strategy, while the suffix distinguishes the individual exit. The tags identify the client-side enrichment to invoke without specifying its visual representation, thereby preserving the separation between marker generation and presentation.

On the client side, we modified a version of the open-source LLM client LibreChat\footnote{LibreChat repository: \url{https://github.com/danny-avila/LibreChat}.} to recognize and handle these three client exits. The modified client interprets the markers embedded in a response and integrates the corresponding supplementary information into the displayed answer, rather than presenting the markers as ordinary text. The client remains responsible for the concrete presentation of each enrichment.

After connecting the modified LibreChat instance to the fine-tuned model, the prototype successfully integrated additional, non-textual information into the LLM's displayed answers. This demonstrates the end-to-end operation of the client-exit mechanism: the model supplies markers within its response, and the client turns those markers into enrichments within the answer.

\begin{figure}[htb]
    \centering
    \includegraphics[width=0.7\textwidth]{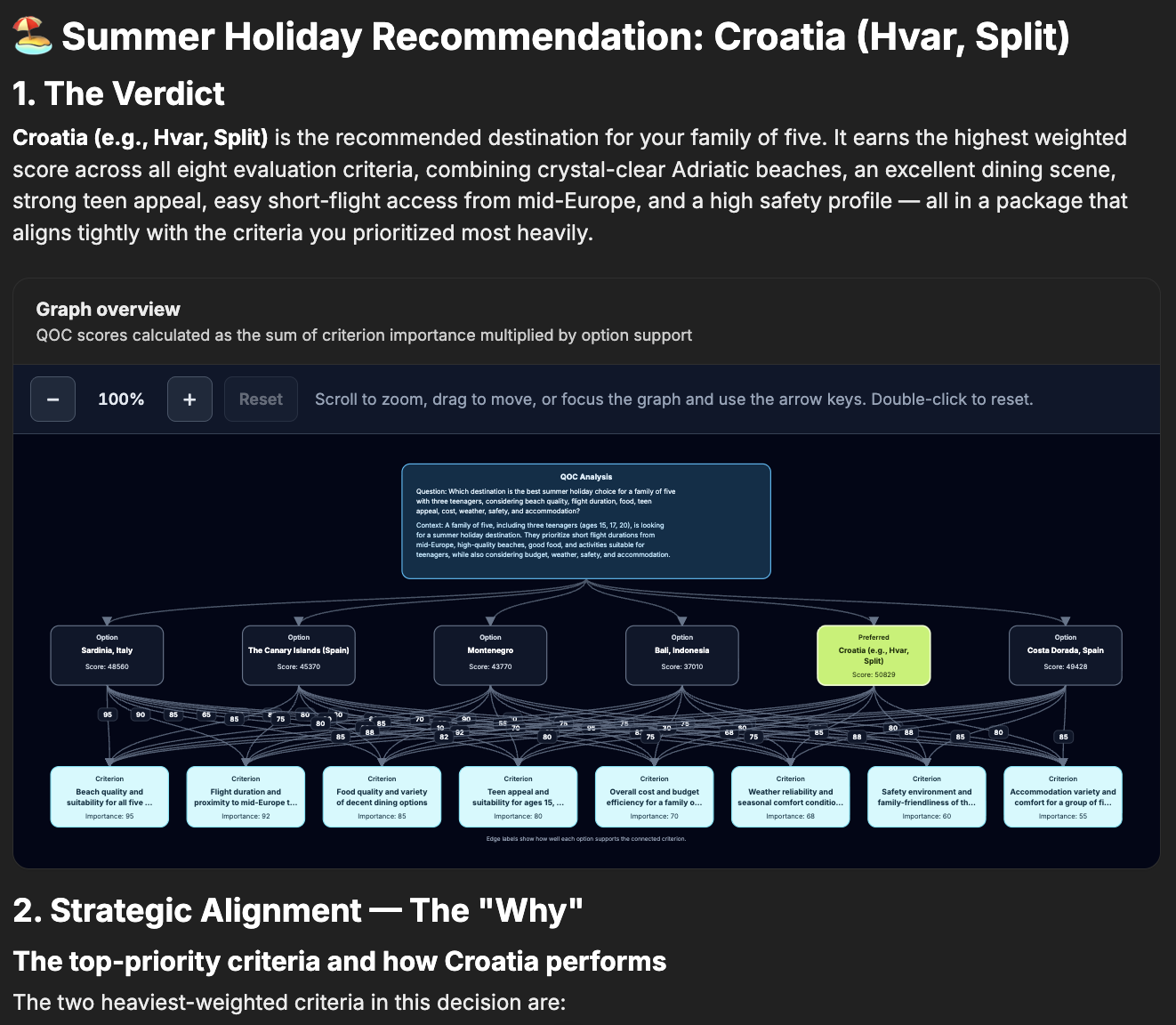}
    \caption{Graph overview integrated into the LibreChat response through the \texttt{qoc.GraphOverview} client exit.}
    \label{fig:implementation-graph-overview}
\end{figure}

Figure~\ref{fig:implementation-graph-overview} shows the graphical representation generated by the \texttt{qoc.Graph\-Overview} client exit for the example question, ``Where should we spend our next summer holiday?'' Following the QOC paradigm, the graph connects the question to the available destination options and relates each option to the criteria used in the analysis, such as beach quality, flight duration, and overall cost. The criterion nodes display their importance, while the edges between options and criteria indicate how strongly each option supports the corresponding criterion. Each option is annotated with an aggregate score, calculated as the sum of its criterion support values weighted by the respective criterion importance. The preferred option, Croatia (Hvar, Split), is visually highlighted. By embedding this graph within the textual response, the client makes the relationships between the question, options, and criteria transparent for inspection alongside the recommendation.
 
\begin{figure}[htb]
    \centering
    \includegraphics[width=0.7\textwidth]{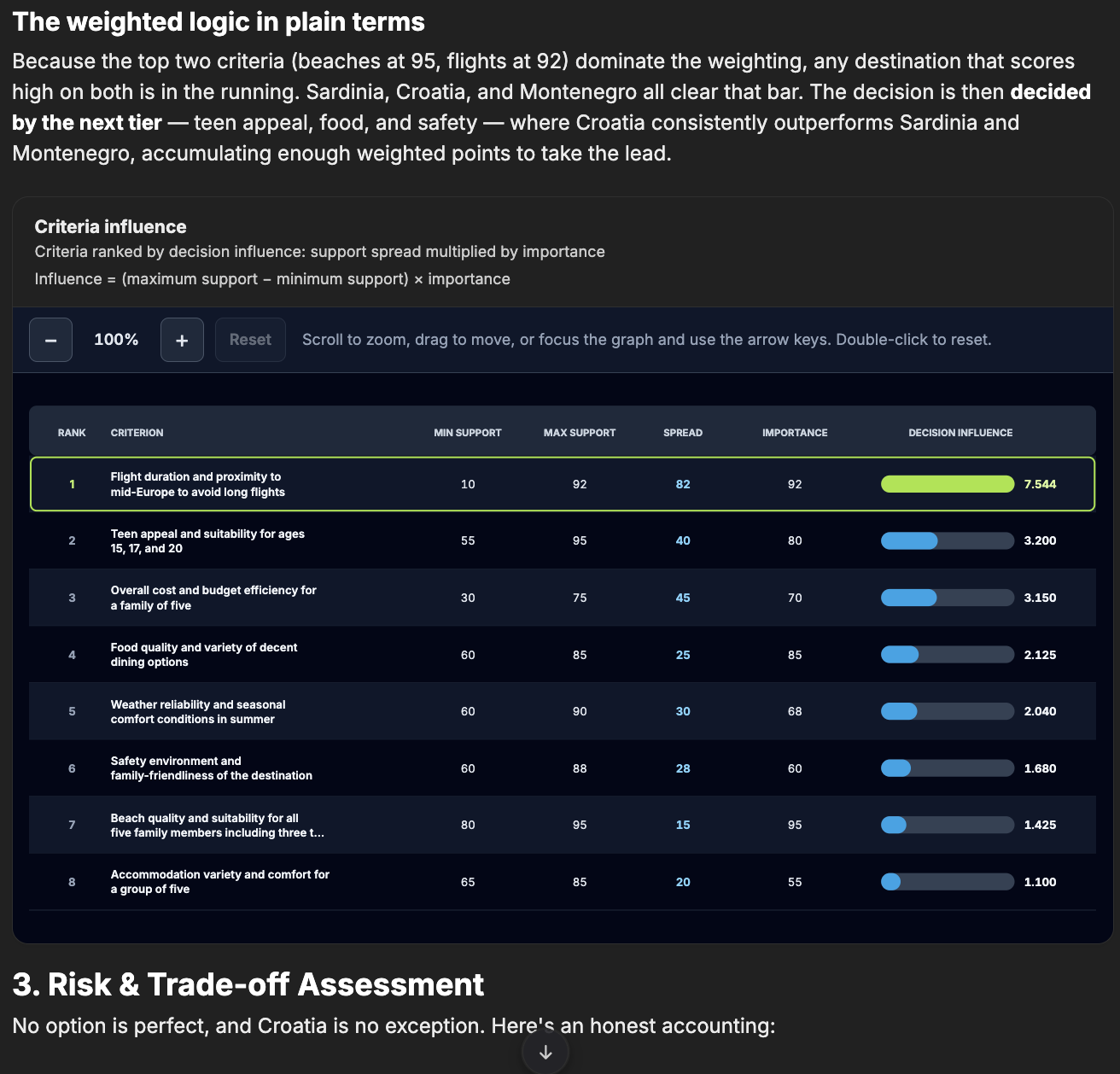}
    \caption{Criteria influence displayed through the \texttt{qoc.CriteriaInfluence} client exit, ranking criteria by their support spread multiplied by their importance.}
    \label{fig:implementation-criteria-influence}
\end{figure}

Figure~\ref{fig:implementation-criteria-influence} shows the criteria influence view generated by the \texttt{qoc.Criteria\-Influence} client exit. The criteria are ranked in descending order of their impact values, calculated by multiplying each criterion's support spread by its importance. The spread is the difference between the maximum and minimum support values across the available options. This measure therefore captures how strongly a criterion differentiates the options after accounting for its assigned importance, rather than reflecting importance alone. The table displays the underlying values alongside horizontal bars that visualize the resulting impact scores. In the holiday example, flight duration and proximity to mid-Europe have the highest impact value, followed by teen appeal and overall cost. This representation allows users to inspect the relative discriminating influence of the criteria underlying the recommendation; it does not, by itself, establish which criterion determines the winning option.

\begin{figure}[htb]
    \centering
    \includegraphics[width=0.7\textwidth]{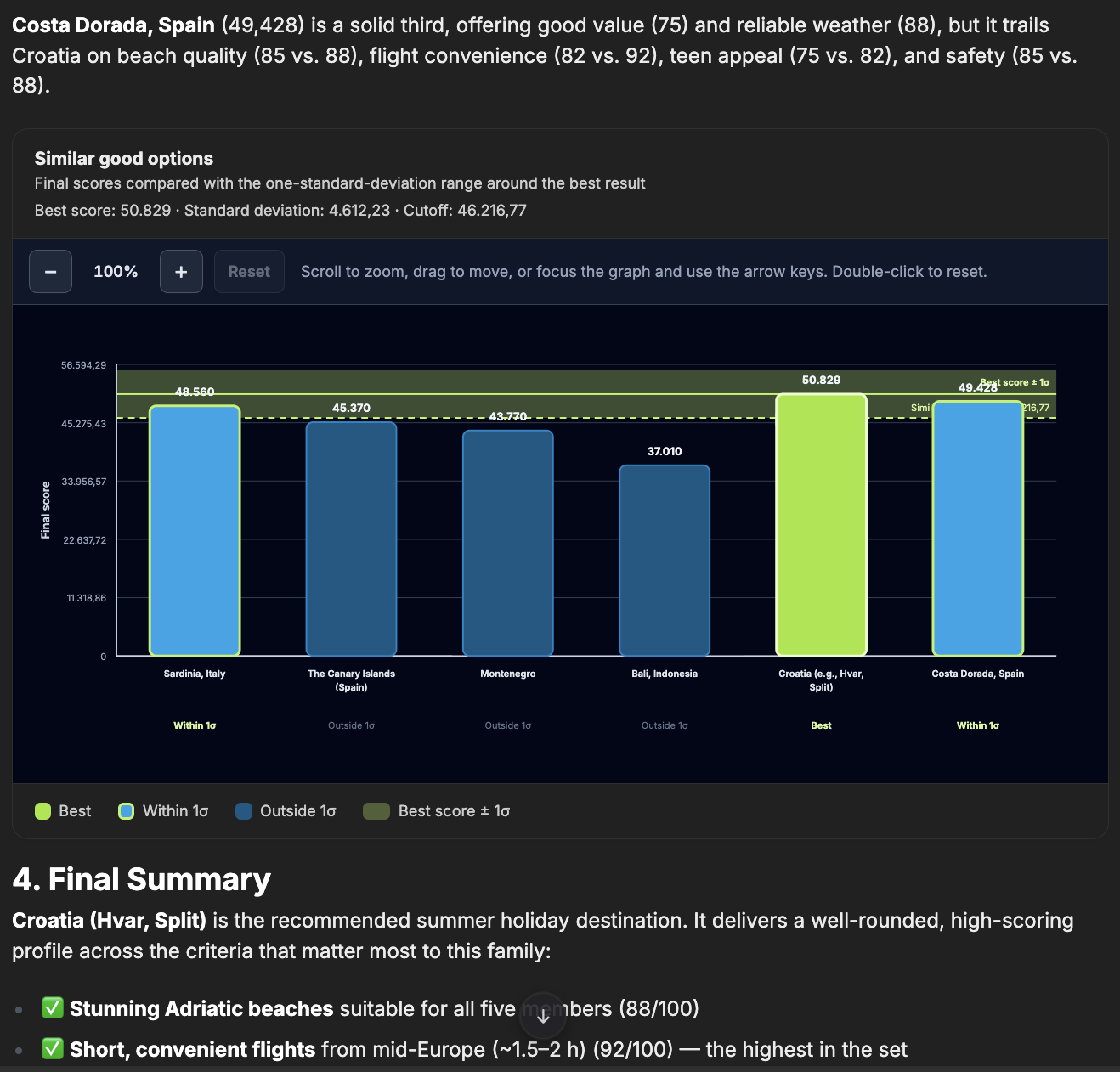}
    \caption{Comparison of option scores through the \texttt{qoc.SimilarGoodOptions} client exit, highlighting alternatives within one standard deviation of the best score.}
    \label{fig:implementation-similar-good-options}
\end{figure}

Figure~\ref{fig:implementation-similar-good-options} shows the statistical comparison generated by the \texttt{qoc.Similar\-GoodOptions} client exit. The visualization compares the aggregate scores of all evaluated options and highlights alternatives considered similarly well suited to the winning option. In this prototype, an alternative qualifies if its score lies within one standard deviation of the highest score, where the standard deviation is calculated across the aggregate scores of all evaluated options. Since no option exceeds the winning score, the effective cutoff is the highest score minus this standard deviation. The bar chart distinguishes the winning option, alternatives meeting the cutoff, and options outside this range. In the holiday example, Sardinia and Costa Dorada qualify as similarly well-suited alternatives to Croatia. This provides users with a view of competitive alternatives rather than presenting only a single recommendation. The one-standard-deviation rule is an operational definition of comparable suitability, not a statistical test establishing equivalence between the options.

%% file: sections/evaluation/evaluation.tex
\section{Evaluation}
\label{sec:evaluation}
To evaluate whether responses containing client exits were preferred over responses without them, we compared two finetuned versions of the
Qwen3.8-27B base model. Both versions were trained to structure their answers according to the Question--Option--Criteria (QOC) model. One version was additionally trained to insert the client exits described in section~\ref{sec:implementation}, whereas the other was trained without these markers. The two finetuning datasets were based on essentially the same examples, with the client-exit version enriched with the corresponding markers.

This setup enabled a comparison between two QOC-oriented models differing in their training for client-exit generation.

We used an LLM-as-a-Judge \cite{zheng2023judging} approach to compare the models' responses to a test dataset comprising 100 questions suited to QOC-based analysis. 

In all 100 test cases, the model finetuned to include client exits incorporated all three aforementioned client exits into every response, corresponding to a complete marker-inclusion rate of 100\%. This result demonstrates that the base model successfully learned through finetuning to generate the specified client-exit markers consistently on the evaluated test set.

Each question was submitted to both fine-tuned models, and the resulting pair of answers was
presented to the judge for a preference assessment. Before making these assessments, the judge received a theoretical explanation of client exits and their intended function. This explanation did not describe client exits as inherently beneficial or instruct the judge to favor responses containing them. The judge was then asked which answer it preferred.

The judge preferred the response from the model trained to include client exits in 66 of the 100 comparisons, corresponding to a preference rate of 66\%. Thus, responses from this model were preferred in a majority of the evaluated cases. This result provides initial evidence that, in the described pairwise assessment, the inclusion of client exits was compatible with answers that the judge found preferable to those produced by the QOC-only variant.

The result should be interpreted as a preference assessment of generated answers, rather than as a direct evaluation of the client-side visualizations or their effects on users. In particular, it does not establish improvements in factual accuracy, user understanding, or decision quality. 

%% file: sections/conclusion-and-future-work/conclusion-and-future-work.tex
\section{Discussion, Conclusion, and Future Work}
\label{sec:conclusion-and-future-work}
The central contribution of Client Exits is the separation of an enrichment's semantic purpose from its client-specific realization. The model identifies a point at which supplementary information may be useful, while the client retains authority over its interpretation and presentation. The prototype demonstrates this division of responsibility through three QOC-specific enrichments integrated into LibreChat. Its significance is not limited to inserting graphics into an answer: the same architectural principle permits different clients to realize an intent differently without requiring the response to prescribe a particular interface. This broader portability remains an architectural possibility, however, rather than an empirically established result of the present single-client implementation.

The evaluation provides evidence for two distinct findings. First, the client-exit model included all three markers in every response to the 100 test questions. This establishes that the specified marker-generation behavior was successfully learned under the tested conditions. It does not establish that the model can selectively choose the appropriate exits, place them optimally, or omit them when enrichment is unnecessary. Indeed, consistently emitting all three markers is compatible with learning a fixed output pattern. A more demanding evaluation must distinguish syntactic inclusion from context-sensitive use.

Second, the LLM judge preferred the client-exit model's responses in 66\% of the comparisons. This is a promising initial preference result, but not a direct measure of the effectiveness of the rendered enrichments. Because the comparison involved separately fine-tuned models, differences in answer content or structure may also have influenced the outcome. Explaining the mechanism without endorsing it avoids an explicit instruction to favor Client Exits, but does not eliminate possible judge biases. The reported experiment therefore supports a preference for the answers produced by one model variant, rather than isolating a causal benefit of the markers themselves. Its scope is further limited to one base model, one thinking strategy, three exit types, and the evaluated question set.

The meaning of transparency also requires qualification. The graph and statistical views expose the structure of the generated QOC analysis, not the internal computations that produced it. Making a rationale inspectable does not guarantee that its criteria, support values, or recommendation are correct. In particular, the criteria-influence measure describes weighted differences across options rather than proving which criterion determines the winner. Similarly, the one-standard-deviation rule defines comparable suitability operationally; it is not evidence of statistical equivalence or a confidence interval. These distinctions should remain visible in the interface so that additional structure does not imply stronger evidential support than the underlying analysis provides.

A first priority is a controlled user study comparing plain answers with client-enriched presentations of the same underlying content. Holding the answer and QOC data constant would help isolate the contribution of rendering from differences in text generation. Relevant outcomes include users' ability to identify decisive trade-offs, detect inconsistencies, compare alternatives, and explain their decisions, alongside completion time and perceived workload. Preference and perceived transparency should be measured separately from these task-based outcomes. Graphical, textual, and voice-based realizations could then test whether the same semantic exits remain useful across interaction modalities and accessibility needs.

A second priority is a more discriminating evaluation of marker generation. Test cases should include questions requiring all, some, or none of the available enrichments, with explicit assessments of marker identity, placement, and unnecessary repetition. Experiments across base models, thinking strategies, and task domains would examine generalization beyond the current setup. Comparisons between prompting and fine-tuning would clarify when additional training is necessary. For preference evaluation, counterbalancing answer order, using multiple judges, and reporting ties, repeated judgments, and uncertainty would strengthen the evidence. Ablations that remove markers from otherwise identical responses would further separate reactions to marker presence from preferences for the generated content.

A third direction concerns the shared contract between model and client. The current namespaced identifiers specify which enrichment is intended, but a reusable implementation also needs an explicit definition of the data required by each handler and how those data are associated with the relevant response segment. Future work should formalize these bindings, namespace versioning, and behavior for unsupported, malformed, or incomplete markers, including responses received incrementally. Capability negotiation could allow clients to advertise supported intents while preserving the model's independence from concrete interface layouts. Evaluation should also measure parsing reliability, enrichment latency, and the usability of fallback presentations.

Finally, client-side control should include control over validation and presentation policy. Future implementations should restrict markers to recognized handlers rather than treating generated content as executable instructions, and should define what happens when required data cannot be validated. For the QOC prototype, tracing displayed values to their source and distinguishing model-generated assessments from client-computed statistics would make the enrichments easier to audit. Sensitivity analyses that vary criterion weights and the similarity threshold could additionally reveal whether the recommendation remains stable under alternative assumptions. These extensions would turn static explanations into opportunities for users to examine and challenge the analysis.

Client Exits transfer the extension-point principle to the presentation boundary of LLM systems. The implemented prototype demonstrates client-controlled enrichment, while the evaluation establishes complete marker inclusion on the 100-question test set and a 66\% judge preference for the client-exit model's answers. Together, these findings support the feasibility of the mechanism without establishing improved human understanding or decision quality. The broader contribution is an architecture in which models indicate opportunities for enrichment and clients determine their realization, providing a basis for more inspectable and adaptable LLM interactions.